\documentclass[10pt,twocolumn,letterpaper]{article}
\pdfoutput=1
\usepackage{comment}
\usepackage{epstopdf}
\usepackage{cvpr}
\usepackage{times}
\usepackage{graphicx}
\usepackage{amsmath}
\usepackage{amssymb}
\usepackage{mathtools}
\usepackage{cuted}
\usepackage[export]{adjustbox}

\DeclareMathOperator*{\argmin}{argmin}

\cvprfinalcopy 

\def\cvprPaperID{3273} 

\begin{document}

\title{3D Human Pose Estimation = 2D Pose Estimation + Matching}

\author{Ching-Hang Chen\\
Carnegie Mellon University\\
{\tt\small chinghac@andrew.cmu.edu}
\and
Deva Ramanan\\
Carnegie Mellon University\\
{\tt\small deva@cs.cmu.edu}
}

\maketitle

\begin{abstract}
We explore 3D human pose estimation from a single RGB image. While many approaches try to directly predict 3D pose from image measurements, we explore a simple architecture that reasons through intermediate 2D pose predictions. Our approach is based on two key observations (1) Deep neural nets have revolutionized 2D pose estimation, producing accurate 2D predictions even for poses with self-occlusions (2) "Big-data"sets of 3D mocap data are now readily available, making it tempting to ``lift" predicted 2D poses to 3D through simple memorization (e.g., nearest neighbors). The resulting architecture is straightforward to implement with off-the-shelf 2D pose estimation systems and 3D mocap libraries. Importantly, we demonstrate that such methods outperform almost all state-of-the-art 3D pose estimation systems, most of which directly try to regress 3D pose from 2D measurements.
\end{abstract}

\section{Introduction}
Inferring 3D human pose from image measurements is classic task in computer vision, dating back to the iconic work of Hogg~\cite{hogg1983model} and O'Rourke and Badler~\cite{o1980model}. Such a technology has immediate applications in various tasks such as action understanding, surveillance, human-robot interaction, and motion capture, to name a few. As such, it has a long and storied history. We refer the reader to various surveys for a broad overview of the popular topic~\cite{forsyth2006computational,moeslund2001survey}.

\begin{figure}[t!]
\centering
\includegraphics[height=3.6in]{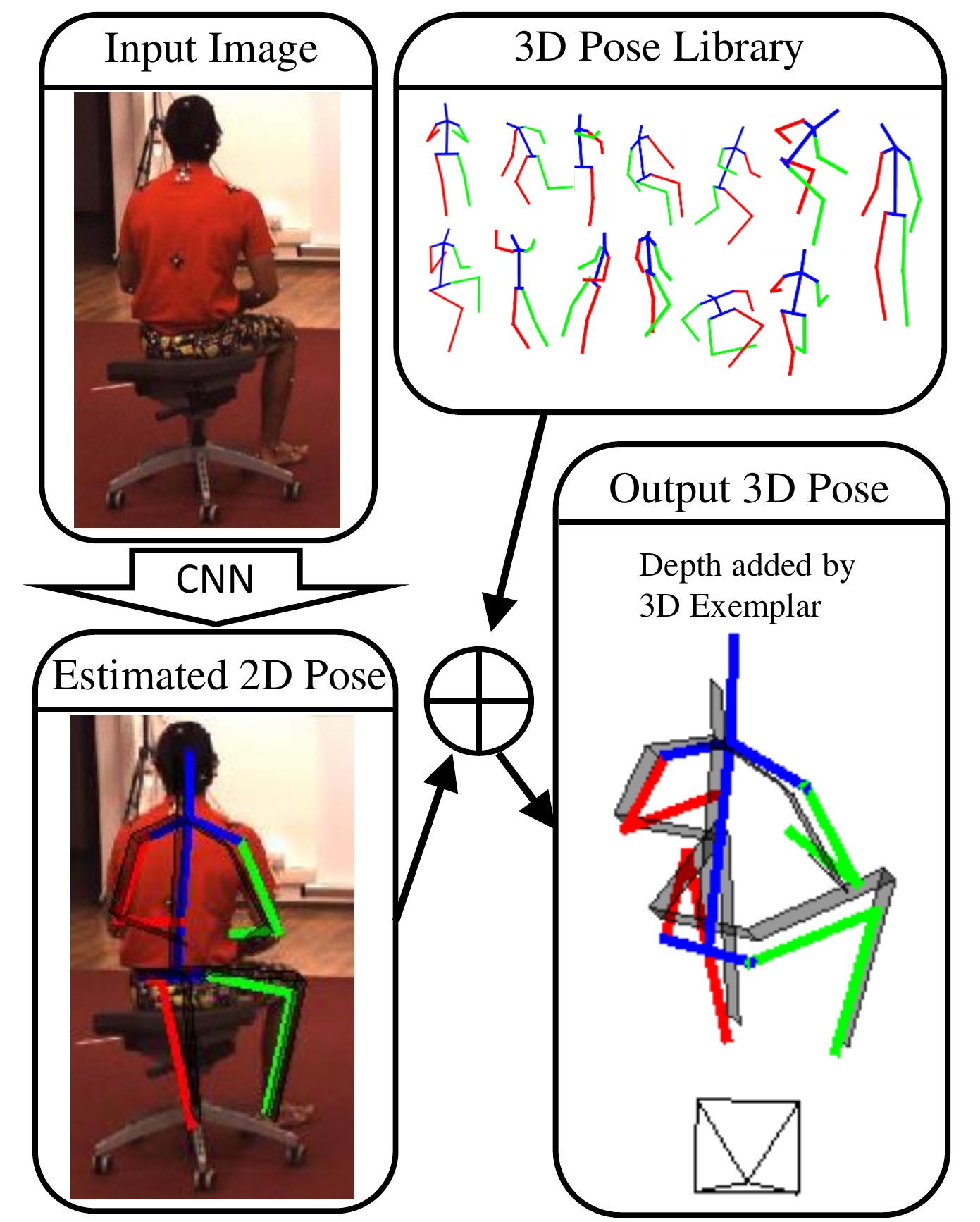}
   \caption{Overview of our approach for 3D pose estimation: given an input image, first estimate a 2D pose and then estimate its depth by matching to a library of 3D poses.  The final prediction is given by the colored skeleton, while the ground-truth is shown in gray. Our approach works surprisingly well because 2D pose estimation is accurate even during occlusions (as illustrated by both wrists above), suggesting that 2D pose estimates need only be refined by adding depth values.}
\label{fig:overview}
\end{figure}

Previous approaches often make use of a highly sensored environment, including video streams~\cite{Zhou_2016_CVPR,Tekin_2016_CVPR}, multiview cameras~\cite{amin2013multi,hofmann2012multi}, depth images~\cite{rafi2015semantic,yub2015random,shotton2013real}. In this work, we focus on the "pure" and challenging setting of recovering 3D body pose with a single 2D RGB image~\cite{li20143d,Yasin_2016_CVPR,rogez2016mocap,wang2014robust}.

Our key insight to the problem is leveraging recent advances in 2D image understanding, made possible through the undeniable impact of deep learning. While originally explored for coarse recognition tasks such as image classification, recent methods have extended such network architectures to ``fine-grained" human pose estimation, where the task is formulated as one of {\em 2D heatmap prediction}~\cite{Wei_2016_CVPR,newell2016stacked,tompson2014joint,hu2015bottom}. One of the long standing challenges in 2D human pose estimation has been estimating poses under self-occlusions. Indeed, reasoning about occlusions has been one of the underlying motivations for working in a 3D coordinate frame rather than 2D. But one of our salient conclusions is that state-of-the-art methods do a surprisingly good job of 2D pose estimation {\em even under occlusion}. Given this observation, the remaining challenge is predicting depth values for the estimated 2D joints.

Inferring 3D structure from 2D correspondences is also a well-studied problem in computer vision, often addressed in multiview setting as structure from motion. In the context of monocular human pose estimation, the relevant cues seem to be semantic rather than geometric. One can estimate 3D postures from a 2D skeleton based on high-level knowledge derived from anthropometric, kinematic, and dynamic constraints. Inspired by the success of data-driven architectures, we explore a simple non-parametric encoding of such high-level constraints: given a 3D pose library, we generate a large number of 2D projections (from virtual camera views). Given this training set of paired (2D,3D) data and predictions from a 2D pose estimation algorithm, we report back the depths from the 3D pose associated with the closest matching 2D example from our library. Our entire pipeline is summarized in Fig.~\ref{fig:overview}. 


{\bf Generalization:} One desirable property of our two-stage approach is generalization. Due to the difficulty of annotation in 3D, training datasets with 3D labels are typically collected in a lab environment, while 2D datasets tend to be more diverse. Our two-stage pipeline makes use of different training sets for different stages, resulting a system that can predict 3D poses from ``in-the-wild" images.

{\bf Evaluation:} Though we present qualitative results on in-the-wild-imagery, we also perform an extensive quantitative evaluation of our method on widely benchmarked 3D human-pose datasets, such as Human3.6M \cite{h36m_pami}. We follow standard train/test protocol splits, but our analysis reveals that there has been inconsistent reporting in the literature, both in terms of test sets and evaluation criteria. To make our results as transparent as possible, we report performance for all metrics and splits we could find. One of our surprising findings is the impressive performance of our simple pipeline: we outperform essentially all prior work on all metrics. Our entire pipeline, even given the non-parametric matching step, returns a 3D pose given a 2D image in {\bf under 200ms} (160ms for 2D estimation by a CNN, 26ms for exemplar matching with a training library of  200,000 poses). Finally, to promote future progress, we perform an exhaustive analysis of additional baselines with upper bounds that reveal the continued benefit of working with intermediate 2D representations and data-driven encoding of 3D constraints.



\section{Related work}
Here we review related works on 3D human pose prediction most relevant to our approach.

\textbf{(Deep) Regression:} Most existing work that makes use of deep features tends to formulate the problem as a direct 2D image to 3D pose regression task. Li \textit{et al.} \cite{li20143d} use deep learning to train a regression model to predict 3D pose directly from images. Tekin \textit{et al.}~\cite{Tekin_2016_CVPR} integrate spatio-temporal features via an image sequence to learn regression model for 3D pose mapping. We provide both a theoretical and empirical analysis that suggests that 2D pose may be a useful intermediate representation.

\textbf{Intermediate 2D pose:} Other approaches have explored pipelines that use 2D poses as an intermediate result. Most focus on the second-stage that lifts 2D estimates to 3D.  This is classically treated as a constrained optimization problem who's objective minimizes the 2D reprojection error of an unknown 3D pose and unknown camera~\cite{Zhou_2016_CVPR,wang2014robust,ramakrishna2012reconstructing,akhter2015pose}. The optimization problem is often subject to kinematic constraints~\cite{wei2009modeling,simo2012single}, and sometimes 3D poses are assumed to live a in low-dimensional subspace to better condition the optimization~\cite{Zhou_2016_CVPR}. Such optimization-based approaches could be sensitive to initialization and local minima, and often require expensive constrained solvers. We use data-driven matching, that when combined with a simple closed-form warping algorithm, yields a fast and accurate 3D solution. 

\textbf{Exemplar-based:} Previous work has also explored example-based methods, dating back at least to~\cite{shakhnarovich2003fast}. A central challenge is generalization to novel poses outside the training set. ~\cite{jiang20103d} propose matching upper and lower bodies individually, to allow for novel compositions at test-time. ~\cite{Yasin_2016_CVPR} adapt exemplars to better match image measurements with an energy minimization approach. \cite{rogez2016mocap} synthesize new 2D images with image-based rendering. Other methods also warp 3D exemplars to 2D image descriptors, often based on shape-context~\cite{agarwal20043d,mori2006recovering} or silhouette features~\cite{IonescuSminchisescu11}.
In our work, we show that a modest number of exemplars (200,000), combined with a simple closed-form algorithm for warping a 3D exemplar to exactly project to 2D pose estimates, outperforms more complex methods.



\section{Approach}
In this section, we describe our method for estimating 3D human pose given a single RGB image. We make use of a probabilistic formulation over variables including the image $I$, the 3D pose ${\bf X} \in \mathbb{R}^{N\times3}$, and the 2D pose ${\bf x}\in \mathbb{R}^{N\times2}$, where  $N$ is the number of articulated joints. We write the joint probability as:
\begin{align}
    {p({\bf X}, {\bf x}, I) = p({\bf X}|{\bf x}, I) \cdot p({\bf x}|I)\cdot p(I)}
    \label{eq:standard_p}
\end{align}
\noindent where the above makes no limiting assumptions by itself.

\begin{figure}[t!]
\centering
\includegraphics[width=\linewidth,clip=true,trim=10mm 70mm 10mm 10mm]{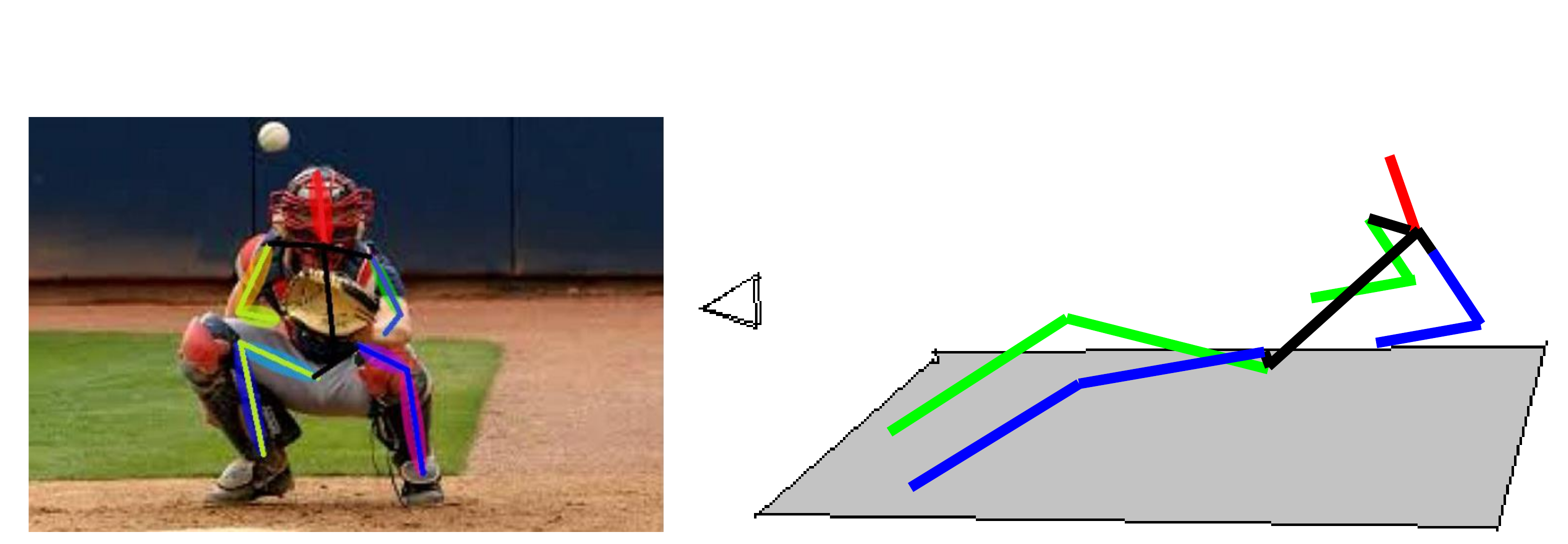}
\caption{A failure case where the 3D pose is {\em not} conditionally independent of the image given the 2D pose: $p({\bf X}|{\bf x},I) \neq p({\bf X}|{\bf x})$. We show the output of our system given the ground-truth 2D pose,  with the (incorrect) best-matching 3D exemplar on the right (visualized from a novel viewpoint, where the estimated camera is drawn as a view frustum). Our experiments suggest that such cases are rare, and that much of the time 3D can be inferred from 2D projections.}
\label{fig:failcase}
\end{figure}

{\bf Conditional independence:} Let us now assume that the 3D pose ${\bf X}$ is {\em conditionally independent} of image $I$ given the 2D pose ${\bf x}$. This is equivalent to the implication that given a 2D skeleton, the prediction of its corresponding 3D skeleton would not be affected by 2D image measurements. While this is not quite true (we show a counter example in Fig.~\ref{fig:failcase}), it seems to be a reasonable first-order approximation. Moreover, this factorization still allows for $p({\bf x}|I)$ to be {\em arbitrarily complex},  which is likely needed to accurately model complex interactions between 2D projections and image features during occlusions. Given this conditional Independence, one can write:
\begin{align}
    p({\bf X}, {\bf x}, I) = \underbrace{p({\bf X}|{\bf x})}_{\text{NN}} \cdot \underbrace{p({\bf x}|I)}_{\text{CNN}} \cdot p(I)
    \label{eq:conditioned_p}
\end{align}
We tackle the second term with a image-based CNN that predicts 2D keypoint heatmaps. 
We tackle the first term with a non-parametric nearest-neighbor (NN) model. We describe each term in turn below.

\subsection{Image-Based 2D Pose Estimation}
Given the above Independence assumption, we would first like to predict 2D pose given image measurements. We model the conditional of 2D pose given an image as
\begin{align}
    P({\bf x}|I) = CNN(I)
\end{align}
\noindent 
where we assume CNN is a nonlinear function that returns $N$ 2D heatmaps (or marginal distributions over the location of individual joints).
We make use of convolutional pose machines (CPMs)~\cite{Wei_2016_CVPR}, which return precisely $N$ heatmaps for individual body joints. We normalize the heatmaps so that they can be interpreted as marginal distributions for each joint. CPM is a near-state-of-the-art pose estimation system ($88.5\%$ PCKh on MPII dataset \cite{andriluka14cvpr}, quite close to the state-of-the-art value of $90.9\%$ ~\cite{newell2016stacked}). Note the off-the-shelf CPM model was trained on MPII dataset, which is a somewhat limited dataset in that annotations are provided through manual inspection. We fine-tune this model on the large scale Human3.6M \cite{h36m_pami} training set, which contains annotations acquired by a mocap system (allowing for larger-scale labeling).


\begin{figure}
\centering
\includegraphics[width=\linewidth,clip=true,trim=0mm 0mm 0mm 0mm]{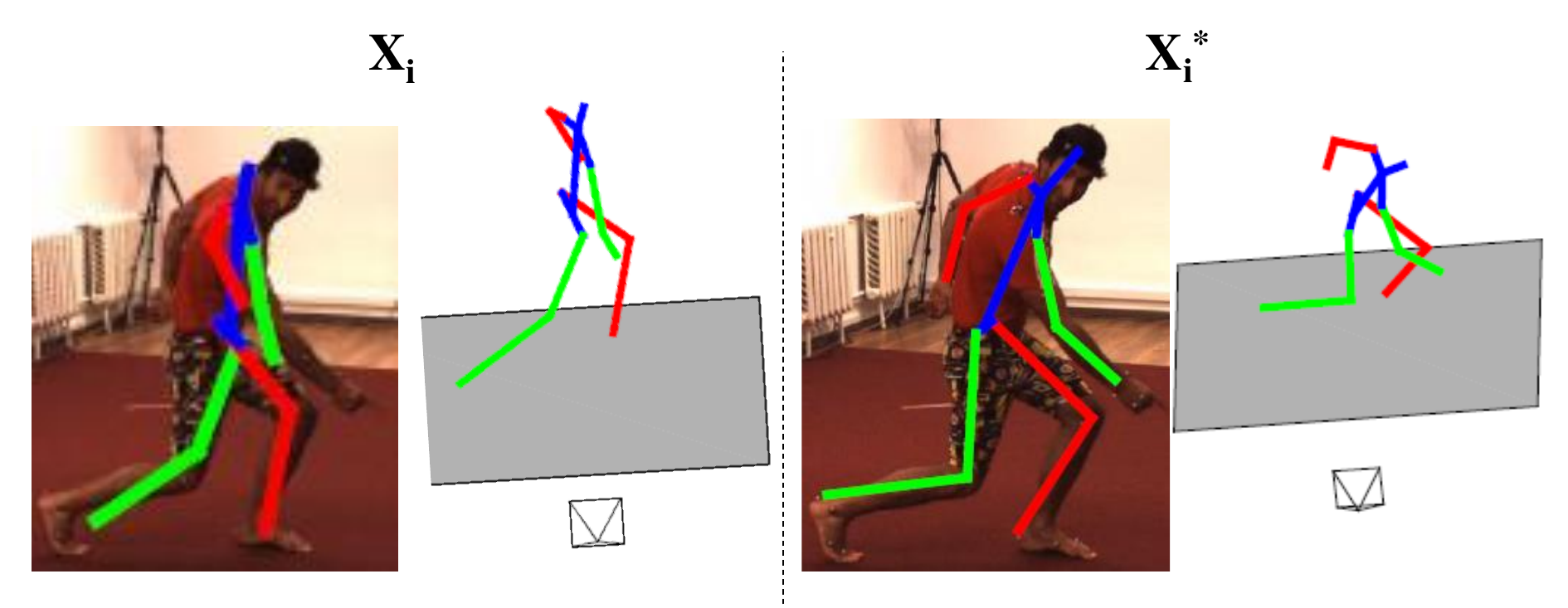}
   \caption{On the left, we show the 3D exemplar ${\bf X_i}$ that best matches the ground-truth 2D pose ${\bf x}$. While the overall pose is roughly correct, the arms and legs are bent incorrectly. By simply copying the depth values from the exemplar (and copying the $(x,y)$ values from the 2D pose under a weak perspective model, as given in \eqref{eq:warp}), we can obtain a warped exemplar ${\bf X_i^*}$ that better matches the 2D pose.}
\label{fig:warp}
\end{figure}

\subsection{Nonparametric 3D shape model} \label{nn}
We model $P({\bf X}|{\bf x})$ with a non-parametric nearest neighbor model. We will follow a notational convention where ${\bf X} = [X,Y,Z]$ and ${\bf x} = [x,y]$. Assume that we have library of 3D poses $\{{\bf X}_i\}$ paired with a particular camera projection matrix $\{M_i\}$, such that the associated 2D poses are given by $\{M_i({\bf X}_i)\}$. If we want to consider multiple cameras for a single 3D pose, we add another copy of the 3D pose with a different camera matrix to our library. We define a distribution over 3D poses based on reprojection error:
\begin{align}
    P({\bf X} = {\bf X}_i|{\bf x}) \propto e^{-\frac{1}{\sigma^2}||M_i({\bf X}_i) - {\bf x}||^2} \label{eq:nn}
\end{align}
\noindent where the MAP estimate is given by the 1-nearest neighbor (1NN).  We explore two extensions to the above basic framework.

{\bf Virtual cameras:} We can further reduce the squared reprojection error by searching over small perturbations of each camera. This involves solving a camera resectioning problem~\cite{hartley2003multiple}, where an iterative solver can be initialized with $M_i$:
\begin{align}
M_i^* = \argmin_M ||M({\bf X}_i) - {\bf x}||^2 \label{eq:view}
\end{align}
In practice, we construct a shortlist of $k$ candidates that score well according to \eqref{eq:nn}, and resort them according to optimal camera matrix. We found that optimizing over cameras produced a small but noticeable improvement in our experiments. Unless otherwise specified, we choose $k=10$ in our experiments.

{\bf Warped exemplars:} Much previous work on exemplars introduce methods for warping exemplars to better match the 2D pose estimates, often formulated as an inverse kinematics optimization problem. We describe an extremely lightweight method for doing so here. We first align the 3D exemplar to the camera-coordinate system used to compute the projection ${\bf x}$. This is done with a 3D rigid transformation given by the camera extrinsics encoded in $M_i$ (or $M^*_i$). In practice we use a training set $\{{\bf X}_i\}$ where 3D exemplars are already aligned to their projections $\{{\bf x}_i\}$, implying that extrinsics in $M_i$ reduce to an identity matrix (which is the case for the Human3.6M dataset~\cite{h36m_pami}, since 3D poses are specified in camera coordinates of their associated image projections). Given this alignment, we simply replace the $(X_i,Y_i)$ exemplar coordinates with their scaled 2D counterparts $(x,y)$ under a weak perspective camera model: \begin{align}
    {\bf X}_i^* = \begin{bmatrix} sx & sy & Z_i \end{bmatrix}, \quad \text{where}\quad  s = \frac{\text{average}(Z_i)}{f} \label{eq:warp}
\end{align}
\noindent where $f$ is the focal length of the camera (given by the intrinsics in $M_i$) and average($Z_i$) is the average depth of the 3D joints. Such weak-perspective approximations are commonly used to initialize algorithms for perspective (PnP) camera calibration~\cite{lu2000fast}, and will be reasonable when the depth variation of the human skeleton is small relative to the overall distance to the camera.  Our results suggest that such closed-form solutions for 3D warping rival the accuracy of complex energy minimization methods (see Fig.~\ref{fig:warp}).

\begin{figure*}[t!]
\includegraphics[width=\linewidth,right]{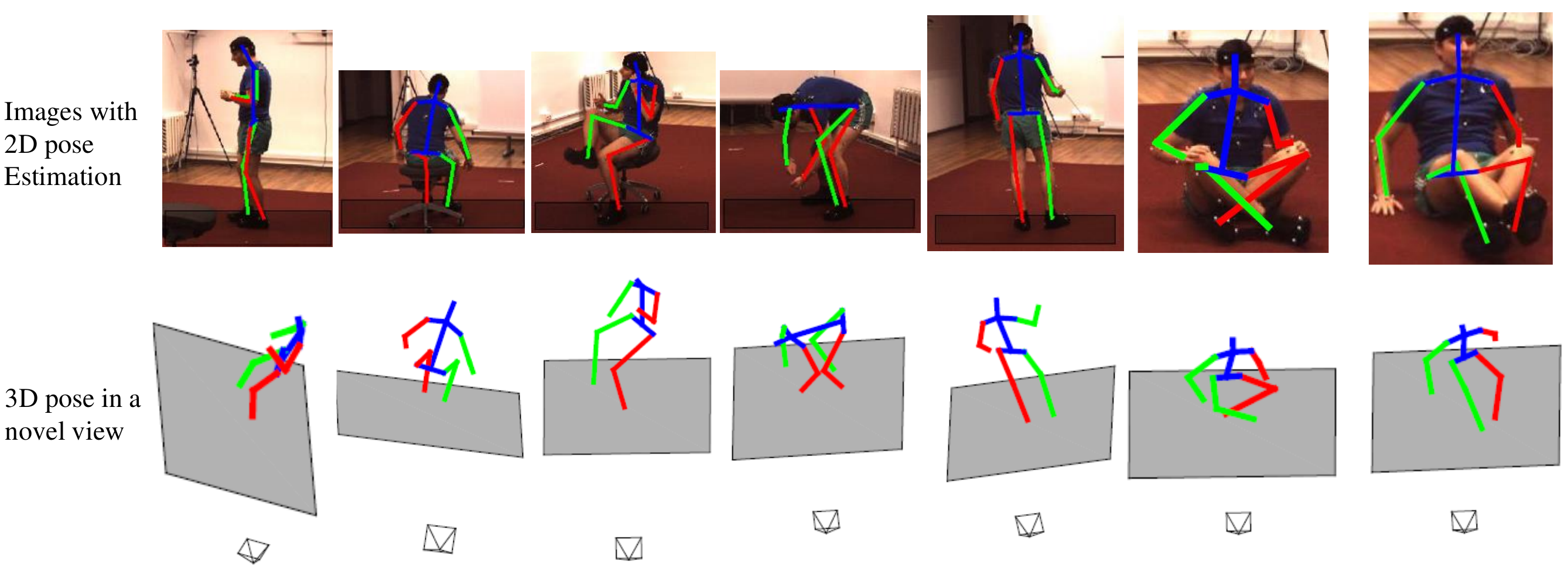}
\includegraphics[width=0.95\linewidth,clip=true,trim=0mm 0mm 0mm 75mm,right]{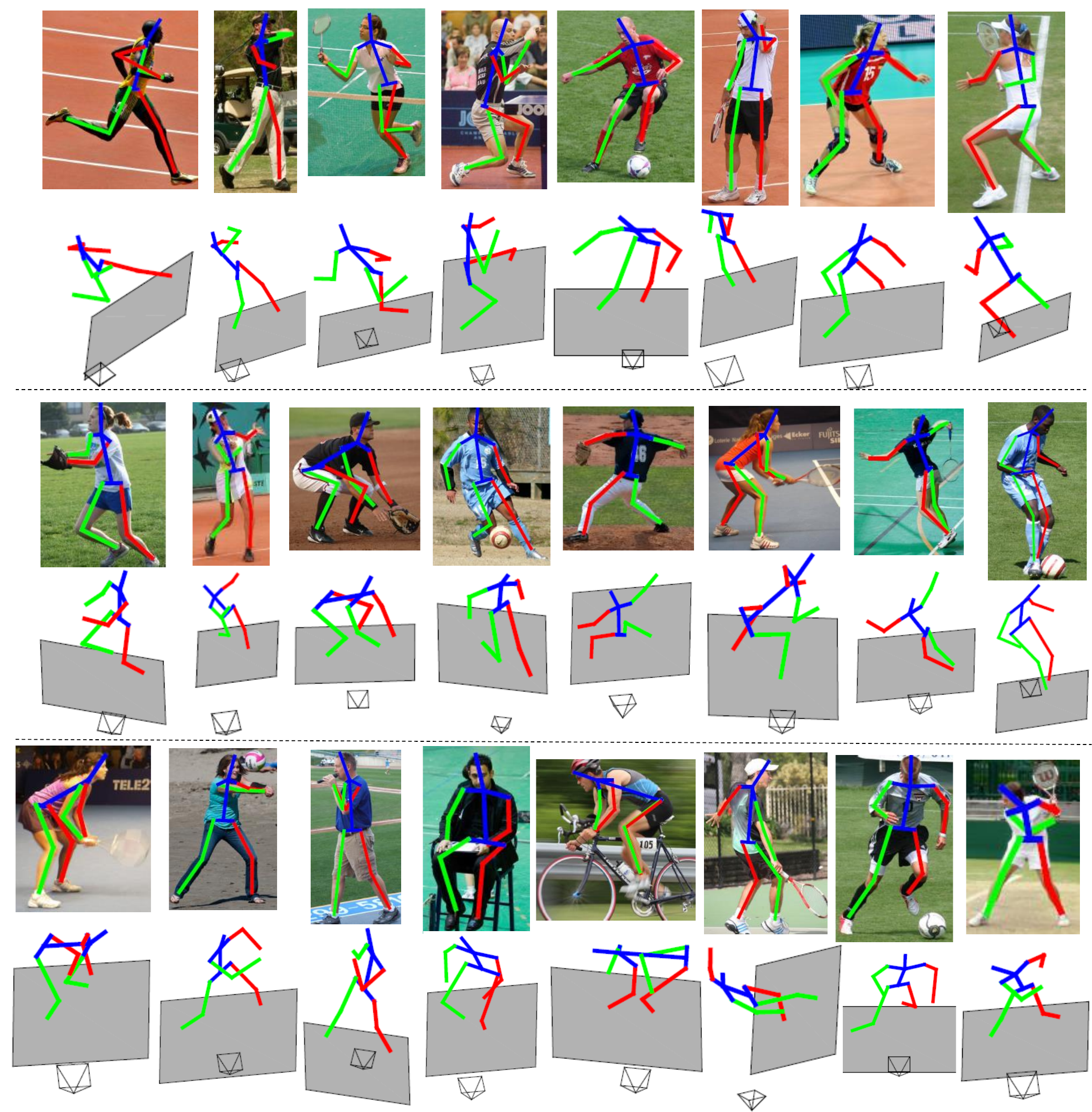}
   \caption{We show qualitative results on Human3.6M-test ({\bf top}) and LSP-test ({\bf bottom}). Our method produces plausible results for challenging images with self-occlusions and extreme poses, and can generalize to activities and poses not in the train set (Human3.6M-train).}
\label{fig:qualitative}
\end{figure*}

\section{Experiments} \label{exp}
In our experiments, we test a variety of variations of our proposed pipeline.

{\bf Qualitative results:} We first present
 some qualitative results. Fig.~\ref{fig:qualitative} shows results on challenging examples from subject S11 of Human3.6M. We choose examples with self-occlusions and sitting poses. To demonstrate the accuracy of the 3D predictions, we visualize novel viewpoints. We then apply the proposed method on Leeds Sports Pose(LSP) dataset \cite{Johnson10} to test cross-dataset generalization.
We posit that our pipeline will generalize across image variation (due to the underlying robustness of our 2D pose estimation system) but maybe limited in the 3D estimates due to the library used (from Human3.6M).
Importantly, our approach produces plausible 3D poses even when the activity class is not included in Human3.6M. This implies that our method can reliably estimate 3D poses in the wild!



\subsection{Evaluation protocols}
We use Human3.6M for quantitative evaluation and analysis. It appears that multiple train/test splits have been used in the literature, as well as different approaches to computing mean per joint position error (MPJPE), measured in millimeters. We summarize them here.

\textbf{Protocol 1:} In~\cite{Yasin_2016_CVPR,kostrikov2014depth,rogez2016mocap}, the entire dataset was partitioned into six training subjects (S1, S5, S6, S7, S8, S9), and one testing subject (S11). Evaluation is performed on every 64th frame of S11's video clips. In this configuration, there are total 1.8 million 3D poses available in the training set. MPJPE between the ground truth 3D pose and the estimated 3D pose is computed by first aligning poses with a rigid transformation~\cite{kostrikov2014depth}.

\textbf{Protocol 2:} Others \cite{Zhou_2016_CVPR,Tekin_2016_CVPR,li20143d} use five subjects (S1, S5, S6, S7, S8)  for training, and two subjects (S9, S11) for testing. We follow \cite{Zhou_2016_CVPR}'s setup that downsamples the videos from $50$ fps to $10$ fps.
Here, MPJPE is evaluated {\em without} a rigid transformation, following the original \textit{h36m} protocol: both the ground-truth and predicted 3D pose is centered with respect to a root joint (\textit{ie.} pelvis). In contrast to Protocol 1, this evaluation can be sensitive to a single poorly-predicted joint, particularly if it is the root~\cite{h36m_pami}.

To compare to published performance numbers, we use the appropriate protocol as needed. From our own experience, we find Protocol 1 to be simpler and more intuitive, and so focus on it for our diagnostic evaluations.


\begin{table*}[t!]
\centering
Mean Per Joint Position Error (MPJPE), in mm\\
\begin{tabular}{|l|c|c|c|c|c|c|c|c|c|}
\hline
Method & Direction & Discuss & Eat & Greet & Phone & Pose & Purchase & Sit & SitDown\\
\hline
\hline
Yasin \cite{Yasin_2016_CVPR}& 88.4 & 72.5 & 108.5 & 110.2 & 97.1 & 81.6 & 107.2 & 119.0 & \textbf{170.8} \\
Rogez \cite{rogez2016mocap}& -&-&-&-&-&-&-&-&-\\
\hline
Ours &\textbf{71.63} &\textbf{66.60} &\textbf{74.74} &\textbf{79.09} &\textbf{70.05} &\textbf{67.56} &\textbf{89.30} &\textbf{90.74} &195.62 \\
\hline
\hline
Method & Smoke & Photo & Wait & Walk & WalkDog & WalkPair & Avg. & Median & -\\
\hline
\hline
Yasin \cite{Yasin_2016_CVPR}& 108.2 & 142.5 & 86.9 & 92.1 & 165.7 & 102.0 & 108.3 & - &-\\
Rogez \cite{rogez2016mocap}&-&-&-&-&-&-& 88.1 & - &-\\
\hline
Ours &\textbf{83.46} &\textbf{93.26} &\textbf{71.15} &\textbf{55.74} &\textbf{85.86} &\textbf{62.51}& \textbf{82.72} & \textbf{69.05}&-\\
\hline
\end{tabular}
\caption{Comparison to \cite{Yasin_2016_CVPR} by \textbf{Protocol 1}. Our results are clearly state-of-the-art. Please see text for more details.}
\label{table: s11_compare}
\end{table*}

\begin{table*}[t!]
\centering
\begin{tabular}{|l|c|c|c|c|c|c|c|c|c|}
\hline
Method & Direction & Discuss & Eat & Greet & Phone & Pose & Purchase & Sit & SitDown\\
\hline
\hline
Yasin \cite{Yasin_2016_CVPR} & 60.0 & 54.7 & 71.6 & 67.5 & 63.8 & 61.9 & 55.7 & 73.9 & 110.8 \\
\hline
${\bf X}^*|gt$ (Ours) &\textbf{53.27} &\textbf{46.75} &\textbf{58.63} &\textbf{61.21} &\textbf{55.98} &\textbf{58.13} &\textbf{48.85} &\textbf{55.60} &\textbf{73.41}\\
\hline
\hline
Method & Smoke & Photo & Wait & Walk & WalkDog & WalkPair & Avg. & Median & -\\
\hline
\hline
Yasin \cite{Yasin_2016_CVPR} & 78.9 & 96.9 & 67.9 & 47.5 & 89.3 & 53.4 & 70.5 & -&- \\
\hline
Ours &\textbf{60.25} &\textbf{76.05} &\textbf{62.19} &\textbf{35.76} &\textbf{61.93} &\textbf{51.08} & \textbf{57.50} & \textbf{51.93}&-\\
\hline
\end{tabular}
\caption{Comparison to \cite{Yasin_2016_CVPR} by \textbf{Protocol 1} given 2D ground truth. Our approach is clearly state-of-the-art, indicating the effectiveness of our simple approach to NN matching and warping. Table~\ref{table: gt2d_nn_comp} shows that even simple NN matching produces an average accuracy of 70.93, rivaling prior art.}
\label{table:s11_compare_gt2d}
\end{table*}

\subsection{Comparison to state-of-the-art (Protocol 1)}

{\bf Final system:} Table~\ref{table: s11_compare} compare MPJPE for each activity class. Our approach clearly outperforms \cite{Yasin_2016_CVPR} and \cite{rogez2016mocap}. ("Ours" in the tables of comparison throughout the experiment refers to the warped exemplar ${\bf X}^*$ described in Section~\ref{nn}) 

{\bf Performance given ground-truth 2D:} A common diagnostic is evaluating performance given ground-truth 2D poses, written as $gt$. Table~\ref{table:s11_compare_gt2d} shows that our simple matching + warping outperforms~\cite{Yasin_2016_CVPR}, who use a complex iterative algorithm for matching and warping exemplars to image evidence. Our diagnostics will later show that even matching exemplars {\em without} warping outperforms prior art, indicating the remarkable power of a simple NN baseline.

{\bf Size of trainset:} Table~\ref{table: train_size_compare} shows the MPJPE versus the training data size. Since approaches deal with 2D and 3D sources differently, we list both sizes. Yasin \textit{et al.} \cite{Yasin_2016_CVPR} project multiple 2D poses from each 3D exemplar (with virtual cameras) to create 2D poses for matching, and Rogez \textit{et al.} \cite{rogez2016mocap} directly synthesize 2D images for training. Our approach makes use of the default training data in Human3.6M, where each 3D pose is paired with a single 2D projection. We max out performance with a modest pose library of $180$k 3D-2D pairs, but produce competitive accuracy even for $18$k. The slight increase in MPJPE for larger training sets seems to be related to noise from 2D pose estimation, since we observe a monotonically decrease when ground truth 2D poses are given (Fig.~\ref{fig:mean_nn}).

\begin{table}[!]
\centering
\begin{tabular}{|l|c|c|c|}
\hline
Method & 2D source & 3D source & Avg. MPJPE\\
\hline
\hline
Yasin \cite{Yasin_2016_CVPR} & 64,000k & 380k & 108.3\\
Rogez \cite{rogez2016mocap}& 207k & 190k & 88.1 \\
\hline
Ours & 18k & 18k & 85.94\\
Ours & 180k & 180k & \textbf{82.37}\\
Ours & 1,800k & 1,800k & 82.72\\
\hline
\end{tabular}
\caption{Comparison to \cite{Yasin_2016_CVPR} and \cite{rogez2016mocap} under different amounts of training data, under {\bf Protocol 1}. Our approach yields the best performance at the source size of $180$k.}
\label{table: train_size_compare}
\end{table}

\begin{table*}[t!]
\centering
\begin{tabular}{|l|c|c|c|c|c|c|c|c|c|}
\hline
Method & Direction & Discuss & Eat & Greet & Phone & Pose & Purchase & Sit & SitDown\\
\hline
\hline
Zhou \cite{Zhou_2016_CVPR}& \textbf{87.36} & 109.31 & \textbf{87.05} & \textbf{103.16} & 116.18 & 106.88 & \textbf{99.78} & \textbf{124.52} & \textbf{199.23} \\
Tekin \cite{Tekin_2016_CVPR}& 102.41 & 147.72 & 88.83 & 125.38 & 118.02 & 112.38 & 129.17 & 138.89 & 224.9 \\
\hline
Ours & 89.87 &\textbf{97.57} &89.98 &107.87 &\textbf{107.31} &\textbf{93.56} &136.09 &133.14 &240.12 \\
\hline
\hline
Method & Smoke & Photo & Wait & Walk & WalkDog & WalkPair & Avg. & Median & -\\
\hline
\hline
Zhou \cite{Zhou_2016_CVPR}& 107.42 & 139.46 & 118.09 & 79.39 & 114.23 & 97.70 & \textbf{113.01} & - &-\\
Tekin \cite{Tekin_2016_CVPR}& 118.42 & 182.73 & 138.75 & \textbf{55.07} & 126.29 & \textbf{65.76} & 124.97 & -&- \\
\hline
Ours & \textbf{106.65} &\textbf{139.17} &\textbf{106.21} &87.03 &\textbf{114.05} &90.55 & 114.18 & 93.05&-\\
\hline

\end{tabular}
\caption{Comparison to \cite{Zhou_2016_CVPR} and \cite{Tekin_2016_CVPR} by \textbf{Protocol 2}. Our results are close to state-of-the-art.}
\label{table: s9s11_compare}
\end{table*}

\subsection{Comparison to state-of-the-art (Protocol 2)}

\begin{figure}[t!]
\centering
Zhou's results \hspace{45pt} Our results
\includegraphics[width=\linewidth,clip=true,trim=0mm 0mm 0mm 8mm]{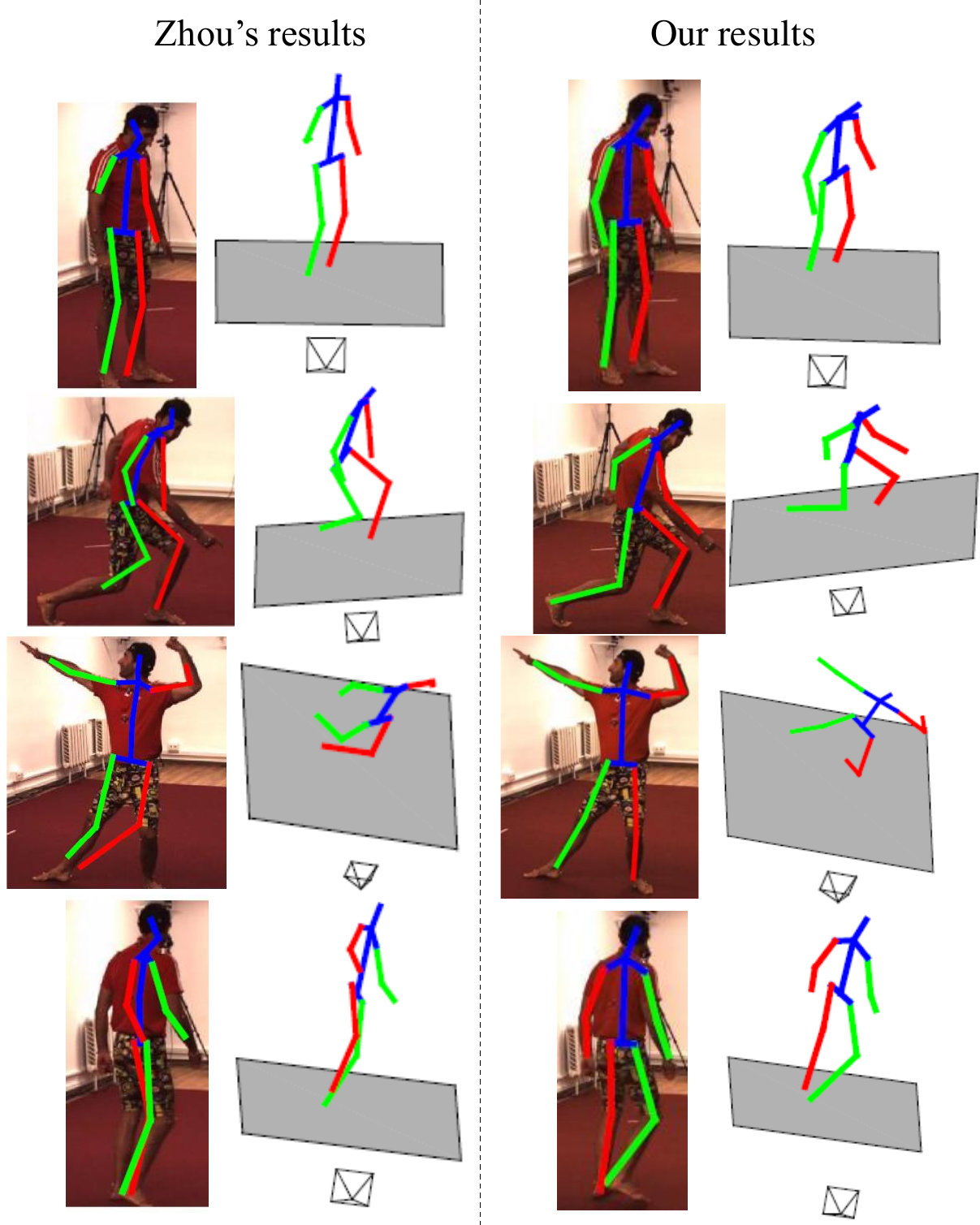}
   \caption{Qualitative comparison of Zhou~\cite{Zhou_2016_CVPR} with our results. 
Our results are generally more accurate, but both methods struggle with left/right limb ambiguities (e.g., the second row). While much of our improved performance comes from better 2D pose estimation, we still compare favorably when using the same ground-truth 2D pose estimates (Fig.~\ref{fig:vis_comp_gt2d} and Table~\ref{table: procrus_compare}).}
\label{fig:vis_comp}
\end{figure}

{\bf Final system:} Table~\ref{table: s9s11_compare} provides the comparison to \cite{Zhou_2016_CVPR} and \cite{Tekin_2016_CVPR} using Protocol 2. Note that in both these works, temporal smoothness was exploited by taking a short image sequences as input. Even though we do not use temporal information, our system is quite close to state-of-the-art. A qualitative comparison to \cite{Zhou_2016_CVPR} is also provided in Fig.~\ref{fig:vis_comp}. 

\begin{figure}[t!]
\centering
Zhou's results \hspace{45pt} Our results
\includegraphics[width=\linewidth,clip=true,trim = 0mm 0mm 0mm 14mm]{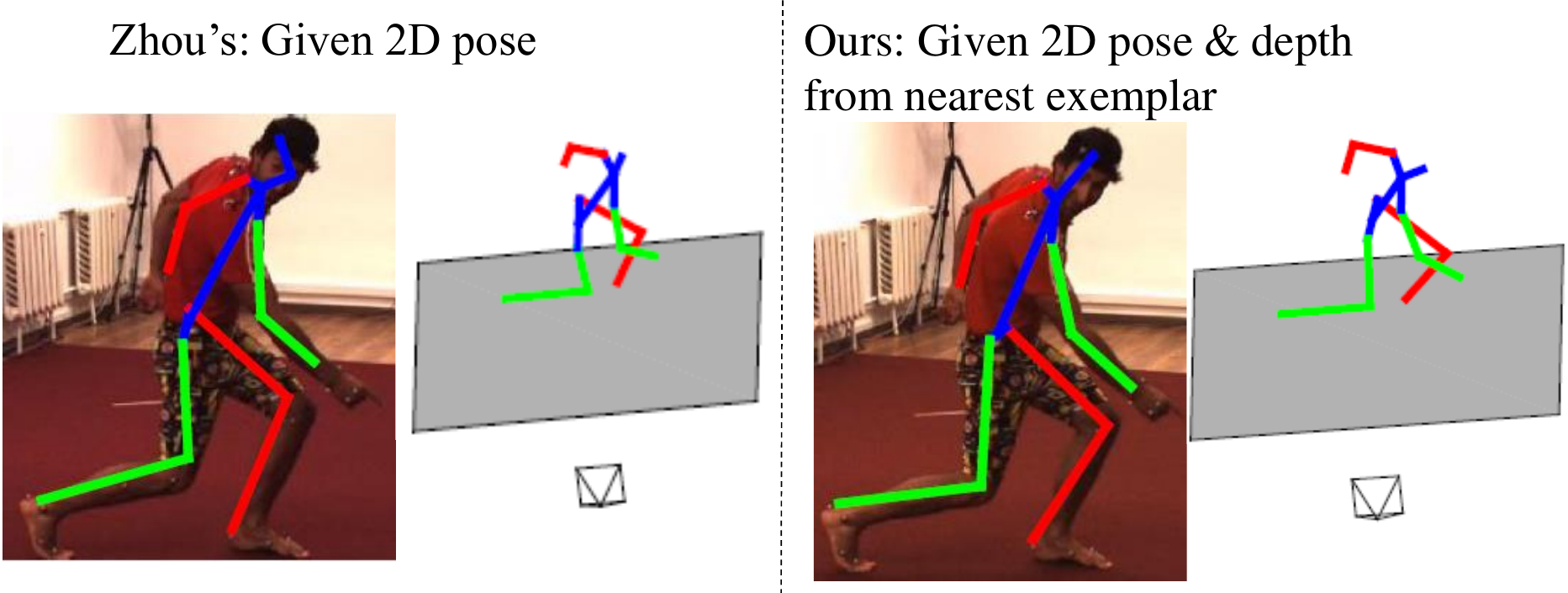}
   \caption{Qualitative comparison of Zhou~\cite{Zhou_2016_CVPR} with our results, given access to the {\em same ground-truth 2D pose}. While both 3D estimates are plausible, Zhou's tends to produce higher 2D reprojection error since 3D poses are restricted to lie in a subspace (e.g., the incorrectly-articulated head).}
\label{fig:vis_comp_gt2d}
\end{figure}

{\bf Performance given ground-truth 2D:}
Our strong performance in Fig.~\ref{fig:vis_comp} might be attributed to better 2D pose estimation. Therefore, we investigate the case given ground truth 2D pose, following Zhou's diagnostic protocol \cite{Zhou_2016_CVPR}: evaluate MPJPE up to a 3D rigid body transformation including scale, only on the first 30 seconds of the first camera in Human3.6M. For a fair comparison, we make use the same training set of 3D-2D training data for both methods. 
The results are shown in Table~\ref{table: procrus_compare}. With a shortlist of $k=10$ matches, camera resectioning \eqref{eq:view} and exemplar warping \eqref{eq:warp} produces a slightly lower error than \cite{Zhou_2016_CVPR}'s approach without a 3D prior.
Qualitative results are provided in Fig.~\ref{fig:vis_comp_gt2d}. Our approach produces lower 2D reprojection error, while Zhou's method appears to suffers from the restriction of 3D poses to a low-dimensional subspace.

\begin{table}[t!]
\centering
\begin{tabular}{|l|c|}
\hline
Method  & Avg. MPJPE \\
\hline
\hline
Ramakrishna~\cite{ramakrishna2012reconstructing}$|gt$, multi-frame & 89.50\\
Dai~\cite{dai2014simple}$|gt$, multi-frame & 72.98\\
Zhou \cite{Zhou_2016_CVPR}$|gt$, single-frame & 50.04 \\
Zhou \cite{Zhou_2016_CVPR}$|gt$, multi-frame & 49.64 \\
\hline
${\bf X}^*|gt$, $k=1$, single-frame & 51.06 \\
${\bf X}^*|gt$, $k=10$, single-frame & {\bf 49.55} \\
\hline
\end{tabular}
\caption{3D pose estimation accuracy given ground-truth 2D poses, under {\bf Protocol 2}. Here, $k$ is the number of candidate exemplar extracted in the shortlist that are subsequently processed by searching over virtual cameras. Our single-frame results with $k=10$ outperforms all prior art, including those which make use of multi-frame temporal cues.}
\label{table: procrus_compare}
\end{table}

\subsection{Diagnostics}
\label{exp1}

We now perform an extensive set of diagnostics to reveal the strength of our individual components, as well as upper-bound analysis that is useful for guiding future work. For simplicity, we restrict ourselves to \textbf{Protocol 1}.

{\bf Effect of warping:} We evaluate the benefits of warping (${\bf X}^*$ vs ${\bf X}$) in Table~\ref{table: cpm_nn_comp}. It is clear that warping exemplars ${\bf X}^*$ is a simple and effective approach to reducing error. Quite surprisingly, {\em even without warping}, simply matching to a set of 3D exemplar projections outperforms the state-of-the-art (see Table 1 \& Table 6)! To analyze an upper-bound for our warping approach, we combine 2D estimates $(x,y)$ with depth values $Z$ given by the ground-truth 3D pose $Z_{GT}$.
In the last row, the performance of combining ground truth depth $Z_{GT}$ with ${\bf x}$ is listed as a reference baseline. This suggests that one can still dramatically lower error by 2X even when continuing to use the output of current 2D pose estimation systems. 

\begin{table}[t!]
\centering
\begin{tabular}{|l|c|c|c|}
\hline
Prediction & Avg. & Median \\
\hline
\hline
${\bf X}|{\bf x}$ & 85.52 & 75.04\\
${\bf X}^*|{\bf x}$ & \textbf{82.72} & \textbf{69.05}\\
\hline
$[s{\bf x}\ Z_{GT}]$ & 43.86 & 30.19\\
\hline
\end{tabular}
\caption{Given the predicted 2D pose ${\bf x}$, warped exemplars ${\bf X}^*$ outperform unwarped exemplars ${\bf X}$ by a reasonable margin. We also compute an upper-bound for warped exemplars that uses $(x,y)$ estimates from the predicted pose and $z$ estimates from the ground-truth 3D pose. The dramatic error reduction suggests that significant further improvement is possible by improving upon our 3D matching. Importantly, this improvement is realizable even given existing 2D pose estimation systems.}
\label{table: cpm_nn_comp}
\end{table}

{\bf Warping given ground-truth 2D:} Next, we compute the error for the case that ground truth 2D pose is given, as shown in Table~\ref{table: gt2d_nn_comp}. We write $|gt$ to emphasize that methods now have access to 2D ground-truth pose estimates. We first note that matching {\em unwarped} examples rivals the accuracy of state-of-the-art (see Table~\ref{table:s11_compare_gt2d} \& Table~\ref{table: gt2d_nn_comp}). This again suggests the remarkable power a simple NN baseline based on matching 2D projections. That said, warping still improves results by a considerable margin. 
A qualitative example is provided in Fig.~\ref{fig:warp}.

\begin{table}[t!]
\centering
\begin{tabular}{|l|c|c|c|}
\hline
Prediction & Avg. & Median \\
\hline
\hline
${\bf X}|gt$ & 70.93 & 65.35\\
${\bf X}^*|gt$ & \textbf{57.50} & \textbf{51.93}\\
\hline
\end{tabular}
\caption{We compare matching to exemplars ${\bf X}$ and warped exemplars ${\bf X}^*$ given ground-truth 2D pose estimates. This suggests that our simple closed-form warping approach would be even more effective with better 2D pose estimates.}
\label{table: gt2d_nn_comp}
\end{table}

{\bf Warping given optimal exemplar match:} It is natural to ask what is the upper-bound on performance given our training set of (3D,2D) pairs. We first compute the optimal exemplar that minimizes 3D reprojection error (up to a rigid body transformation) to the true 3D test pose. We write the index of this best match from the training set as $i=GT$. We would like to see the effect of warping given this optimal match. We analyze this combination in Table~\ref{table: gt3d_nn_comp}. This suggests that, in principle, error can still be significantly reduced (by almost 2X) even given our fixed library of 3D poses. However, it is not clear that this is obtainable given our pipeline because it may require image evidence to select this optimal 3D exemplar (violating the conditional independence assumption from \eqref{eq:conditioned_p}).

\begin{table}[t!]
\centering
\begin{tabular}{|l|c|c|c|}
\hline
Prediction & Avg. & Median \\
\hline
\hline
${\bf X}|GT$ & 60.11 & 55.36\\
${\bf X}^{*}|GT$ & \textbf{37.32} & \textbf{33.91}\\
\hline
\end{tabular}
\caption{We analyze performance given the optimal matching 3D training exemplar "GT" (in terms of 3D error wrt the ground-truth test 3D pose). Simply reporting this optimal match produces an error of 60mm, around 10mm lower than the actual match found given an ideal 2D pose-estimation system (Table 7). Warping this exemplar $X^*|GT$ significantly improves accuracy. This suggests that our overall 3D matching stage could still be significantly improved even given the current size of the library of 3D poses.} 
\label{table: gt3d_nn_comp}
\end{table}


\begin{figure}[t!]
\centering
\includegraphics[width=0.95\linewidth]{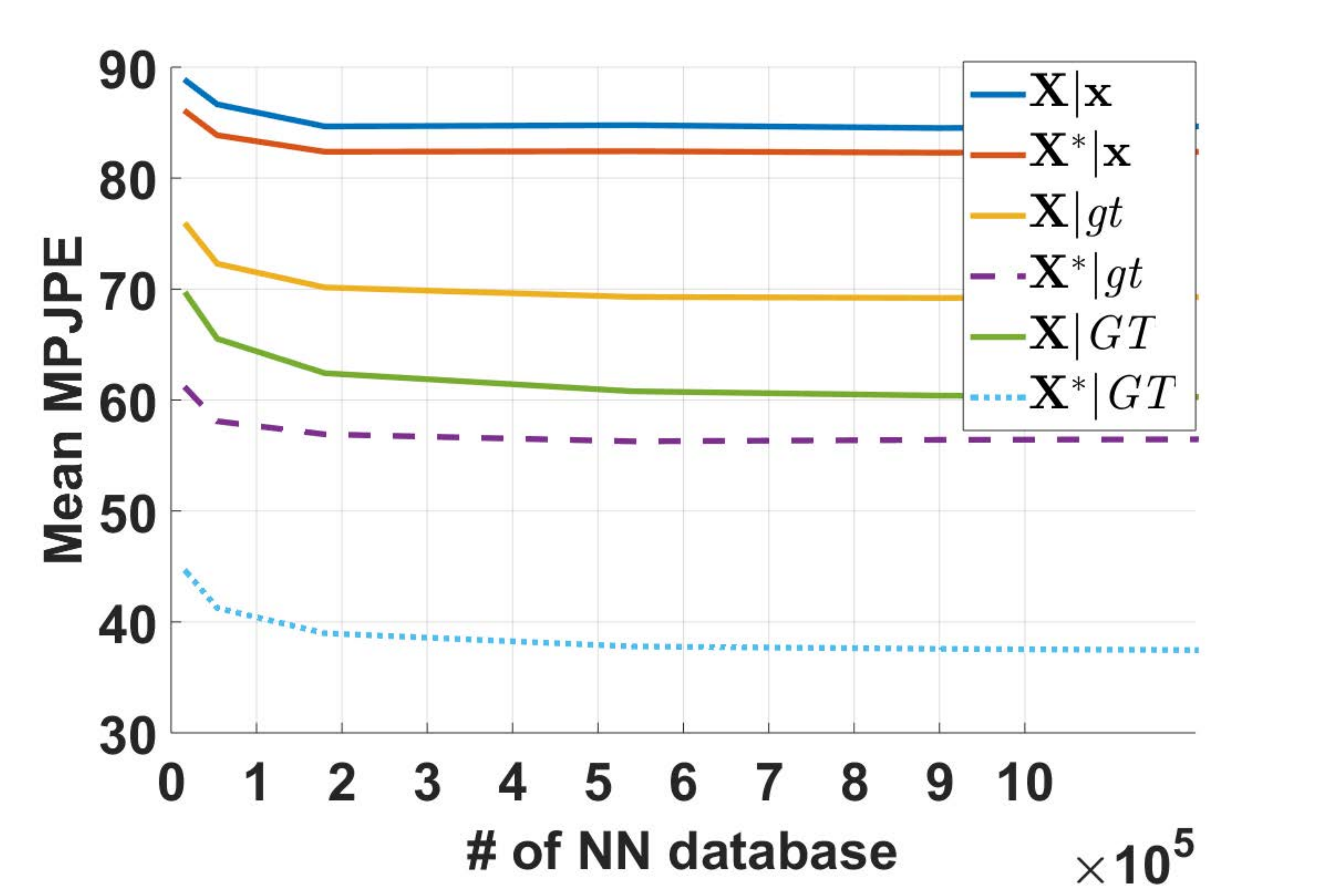}
 \caption{Mean MPJPE by \textbf{Protocol 1} versus size of the 3D pose library. We explore diagnostic variants using previously-introduced notation.
   In general, MPJPE decreases with a larger library. The error saturates at  $2\times10^5$ when using CNN-predicted 2D poses ``$|{\bf x}$", but saturates later at  $5\times10^5$when using ground-truth 2D poses ``$|gt$". The results suggest that with better 2D pose estimates, our exemplar matching would benefit from larger training data.}
 \label{fig:mean_nn}
\end{figure}

\begin{figure}[t!]
\centering
\includegraphics[width=0.95\linewidth]{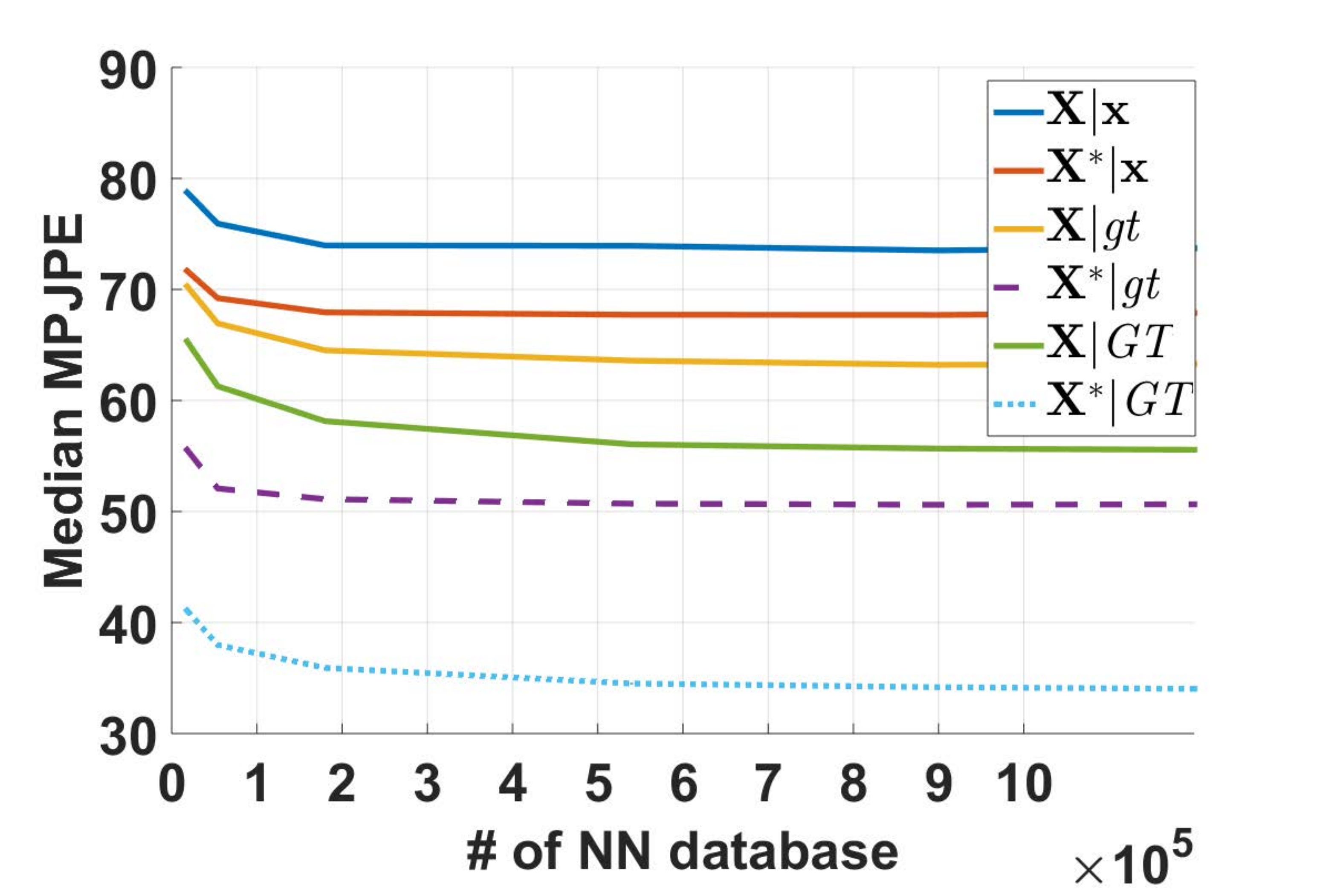}
   \caption{Median MPJPE by \textbf{Protocol 1} versus database size. Median error is lower than mean error in Fig.~\ref{fig:mean_nn}, suggesting that a few joints are responsible for a large mean error. Other trends follow the mean error curves from Fig.~\ref{fig:mean_nn}.}
\label{fig:median_nn}
\end{figure}

{\bf Effect of trainset size:} An important aspect to investigate is the influence of database size. Here we investigate the error versus the number of exemplars in the database. Fig.~\ref{fig:mean_nn} evaluates performance versus a random fraction of our overall database. As expected, more data results in lower error, though diminishing results are observed (even in log scale). This is reasonable since training data is extracted from videos captured at $50$ fps, implying that correlations over frames might limit the benefit of additional frames. We see that convergence is also effected by the quality of the 2D pose estimates: error given ground-truth 2D poses plateaus at $5\times10^5$, while 2D pose estimates plateau even sooner at $2\times10^5$. We posit that a more restricted 3D pose prior (implicitly enforced by a small randomly-sampled 3D library) helps given inaccurate 2D pose estimates. But in either case, {\em exemplar-based 3D matching is effective even for modestly-size training sets (200,000)}. This analysis appears to suggest that better 2D pose estimates are needed to take advantage of ``bigger" 3D datasets.

Since the joint prediction error is not a normal distribution, we also plot median error in Fig.~\ref{fig:median_nn}.
We see that the median is generally lower than mean error, and the difference between the two becomes smaller when ground truth 2D or 3D is given. This may suggest that errors are often due to a single incorrect joint prediction, which would significantly impact average error but not the median.

{\bf Cross-dataset evaluation:} To further examine generalization of exemplar-matching, Table~\ref{table:gt2d_humaneva} quantitatively evaluates accuracy on HumanEva-I \cite{sigal2010humaneva} given a model trained on Human3.6M. These results suggest that the 3D exemplars from HumanEva do generalize, and that generalization is {\em significantly} improved through our warping procedure.

\begin{table}[t!]
\resizebox{\linewidth}{!}{
\begin{tabular}{|l|c|c|c|c|c|c|}
\hline
&Walk & Jog & Throw & Gestures & Box & \bf{Avg.} \\
& & & Catch & & &\\
\hline
\hline
Warped & 64.46 & 69.88 & 59.99 & 67.89 & 79.22 & 68.29\\
\hline
Unwarped & 90.17 & 95.27 & 82.74 & 88.82 & 103.85 & 92.17\\
\hline
\end{tabular}
}
\vspace{5pt}
\caption{We evaluate a Human3.6M-trained model on HumanEva (under Protocol 1).  To isolate the impact of 3D matching, we use ground-truth 2D keypoints. As a point of comparison, average error on Human3.6M test is 70.93 (unwarped) and 57.5 (warped) (Table~\ref{table:gt2d_humaneva}). These results suggest that 3D exemplars do generalize across datasets, and importantly, warping {\em significantly} increases the amount of generalization. Note that the two datasets use different definitions of skeletons, implying that learning a mapping should reduce error even further.}
\label{table:gt2d_humaneva}
\end{table}


\section{Conclusion}
We present an simple approach to 3D human pose estimation by performing 2D pose estimation, followed by 3D exemplar matching. The simplicity and efficiency of our method, combined with its state-of-the-art performance on both benchmark datasets and unconstrained ``in-the-wild" imagery, suggests that such simple baselines should be used for future benchmarking in 3D pose estimation. A notable advantage of intermediate 2D representations is {\em modular training} -- 2D datasets (which are typically larger and more diverse because of ease of annotation) can be used to train the initial-image processing module, while 3D motion capture data can be used to train the subsequent 3D-reasoning module. This allows our system to take immediate advantage of advances in 2D pose estimation, such as multi-body analysis~\cite{eichner2010we}. Our results also suggest that 3D inference is, in some sense, ``all about 2D", at least in the case of articulated objects. Indeed, one of our surprising findings was the high performance of 2D pose estimation systems even under occlusions, suggesting that 2D estimates can in fact be reliably estimated without directly reasoning about depth. Given such reliable 2D estimates, we show that one can efficiently impute depth through simple memorization and warping of a 3D pose library.

{\bf Acknowledgements:} This work was supported by NSF Grant 1618903, NSF Grant 1208598, the Intel Science and Technology Center for Visual Cloud Systems (ISTC-VCS), Google, and Amazon.



\clearpage
{\small
\bibliographystyle{ieee}
\bibliography{egbib}
}

\end{document}